\documentclass[runningheads]{llncs}
\usepackage[T1]{fontenc}
\usepackage{graphicx}
\usepackage{booktabs}
\usepackage{caption}       
\usepackage{subcaption}
\usepackage{multirow}
\usepackage{multirow}
\usepackage{array}
\usepackage{adjustbox}
\usepackage{listings}
\usepackage{enumitem}
\usepackage{tabularx}
\usepackage{float}
\usepackage{makecell}
\usepackage{placeins}
\begin{document}
\title{Evidence-Supported Credit Risk Report Generation Using News-Centric Financial Knowledge Graph}
\titlerunning{Evidence-Supported Credit Risk Report Generation}
%
\author{Rocío Jiménez-Villén\inst{1} \and
Ziwei Xu\inst{2} \and
Ying Chen\inst{2} \and 
Oscar Araque \inst{1} \and
Ryutaro Ichise\inst{2}
}
\authorrunning{R. Jiménez-Villén, Z. Xu et al.}
%
\institute{Universidad Politécnica de Madrid, Spain \and
Institute of Science Tokyo, Japan}
%
\maketitle              
\begin{abstract}
Financial markets evolve in response to real-world events reported in news, yet these drivers often remain implicit in text. To better explain market dynamics, event–market relations must be explicitly modeled through factual, company-centric, and environment-aware knowledge graphs. We present FinKG-News, a framework that automatically constructs such graphs by extracting news events as anchors linked to companies. Using FinKG-News as grounded evidence that integrates events, news, and company data, we develop an in-context learning architecture for credit risk report generation across three core financial dimensions. Automatic and human evaluations show that automated hallucination detection and quality assessment remain unreliable, making expert judgment indispensable. Our approach consistently outperforms baselines, improving quality by 19\%-34\% while reducing hallucinations. The source code and project resources are publicly available at:
\url{https://github.com/ichise-laboratory/FINKG-news}.
\end{abstract}
\section{Introduction}
Financial markets are shaped by real-world events transmitted through news streams and social platforms \cite{manela2017news,baker2016measuring}. Yet these event drivers often stay hidden within textual content. By explicitly modeling event-market relationships, we can provide clearer, fact-based insights for rating, and forecasting financial confidence and market trends. Existing financial knowledge graphs effectively model static corporate relationships like equity ownership~\cite{finkg} and supply chains~\cite{refinitivkg}. However, their lack of real-world dynamic interaction limits their utility for complex financial applications such as macroeconomic forecasting and risk analysis. Moreover, their expert-driven, handcrafted nature creates accessibility barriers for broader usage. In general knowledge applications, domain-specific report generation exhibits severe hallucination issues, particularly in finance due to its diverse user base. We focus specifically on credit risk reporting, which requires knowledge graphs that are company-centric, environmentally interactive, and grounded in factual and traceable information.
\begin{figure*}[t]
    \centering
    \includegraphics[width=1\linewidth]{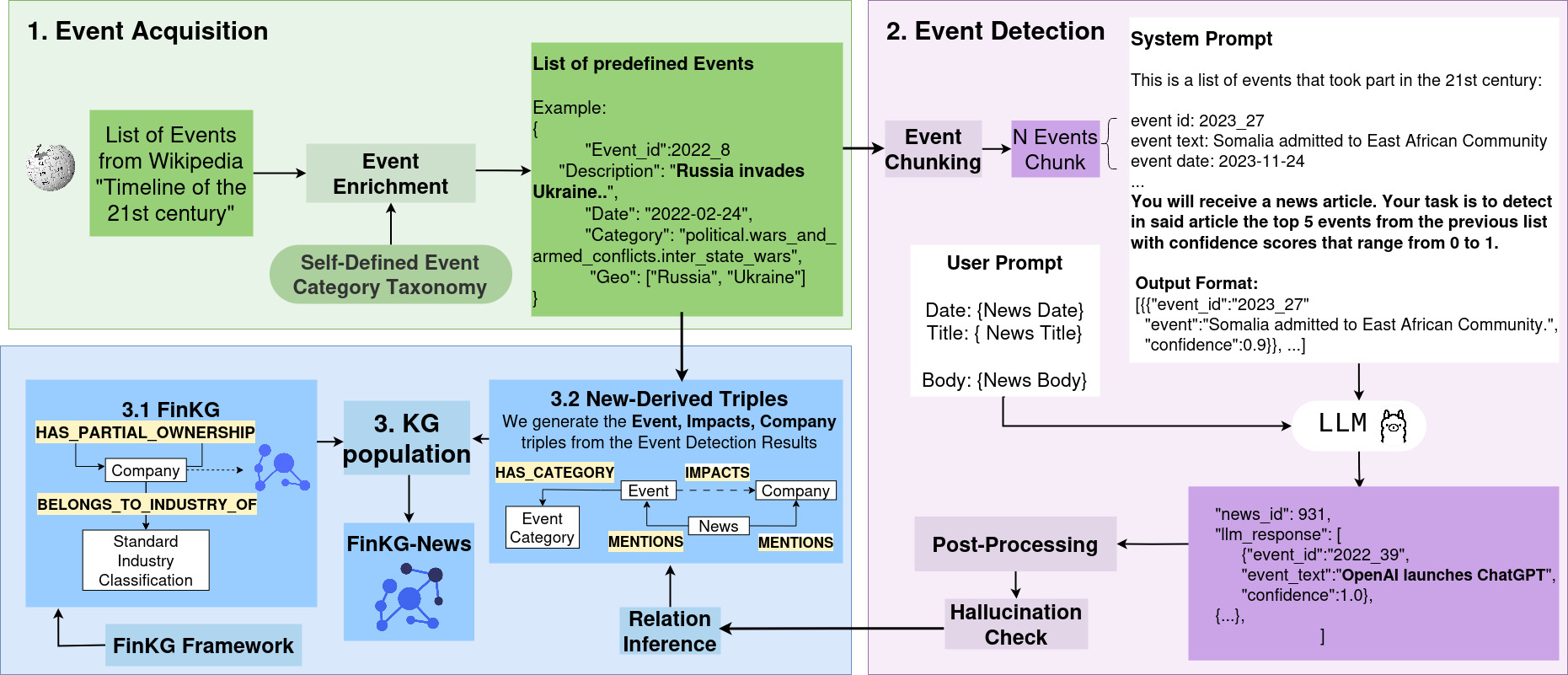}
    \caption{FinKG-News Framework}
    \label{fig_framework}
\end{figure*}
In this work, we propose a framework to automatically generate such knowledge graphs with real-world interaction by detecting events from news as real-world anchors connecting to market companies, which we name \textbf{FinKG-News}. Furthermore, we design a credit risk report generation architecture employing in-context learning techniques to ensemble three different financial dimensions: financial profile, ownership record, and operating environment. Experimental results demonstrate that FinKG-News centralizes 12,047 companies across large, mid, and small capitalization, linking them to 354 real-world events from 2000 to 2025. All identified relationships are leveraged during report generation. Evaluation by both automatic and human assessors reveals that human evaluators exhibit superior performance in hallucination detection compared to automated evaluators. Our generated reports achieve significantly higher content quality with substantially reduced hallucination relative to baseline content, achieving quality gains between 19\% and 34\% across various companies.

The paper starts with an overview of the FinKG-News framework, followed by the credit risk report generation architecture in Section~\ref{sec_generation}. Sections~\ref{sec_exp} and ~\ref{sec_eval} elaborate on experimental setup and evaluation results, while Section~\ref{sec_relatedw} reviews related work. Finally, we conclude and discuss limitations, with supplementary materials provided in the appendix.

\section{FinKG-News Framework}

\subsection{Task Overview}

Figure~\ref{fig_framework} represents the framework followed to construct FinKG-News. We begin by constructing a fixed inventory of historical events. Based on incoming news, we then execute an event detection pipeline comprising several modules: event chunking, prompt engineering, post-processing, and hallucination checking. Using relation inference, the system then extracts and maps event-company relationships. Finally, we utilize the inferred triples and their associated relations to populate FinKG knowledge graph via shared company entities, thereby constructing the complete FinKG-News.

\subsection{Event Acquisition} 
\label{subsec:events}
An event is defined as a historical milestone or incident that is represented by a short description and a specific date or temporal interval. To help distinguish the events in world wide environment, we enrich events with two additional dimension of metadata: 1). their geographical context; 2). event category. Based on a self-defined taxonomy developed inductively from the available events, the first layer categorizes events as political or non-political, with two subsequent layers providing further refinement, as shown in Table \ref{tab:taxonomy} of Appendix \ref{app:taxonomy}. Finally, each event is represented by:   

\noindent
\small
\begin{itemize}[topsep=0pt, partopsep=0pt, itemsep=0pt, parsep=0pt]
  \item \textbf{Identifier}: An event\_id for each unique event.
  \item \textbf{Description}: A concise summary of the event.
  \item \textbf{Date}: The specific date or date range.
  \item \textbf{Category}: Its classification within our self-defined taxonomy.
  \item \textbf{Geo}: Country, continent, or international groupings involved.
\end{itemize}

\subsection{Event Detection}
\label{subsec:event-detection}

In our approach, the task is to identify and retrieve the most relevant events from the list that are related to the given news item. We provide the list of events directly injected in the system prompt of an LLM, along with the task instructions. Then we feed the user prompt with the news article’s date, title, and body (see examples in Block2 of Figure \ref{fig_framework}).
To mitigate hallucinations, each event is attached with a unique identifier, and the model is explicitly asked to output these identifiers along with a confidence score between 0 and 1. Additionally, to reduce the system prompt size, we split the event list into chunks of \(N\) events and prompt the model multiple times for each news article, once for each chunk.

\subsection{Knowledge Graph Population}
FinKG framework~\cite{finkg} can capture fine-grained linkages among companies operating in financial markets; however, it fails to encode their relationships with concrete real-world events. Therefore, this section employs knowledge graph population techniques to integrate news-detected event information into FinKG, forming FinKG-News and enabling more comprehensive knowledge discovery supported by news sources.

\paragraph{FinKG Framework}
To ensure broad company coverage, we use seed firms across different capitalization levels: large-cap\footnote{\scriptsize\texttt{https://en.wikipedia.org/wiki/List\_of\_S\&P\_500\_companies}}, mid-cap\footnote{\scriptsize\texttt{https://en.wikipedia.org/wiki/List\_of\_S\&P\_400\_companies}}, and small-cap\footnote{\scriptsize\texttt{https://en.wikipedia.org/wiki/List\_of\_S\&P\_600\_companies}} companies.
The key entities and relations (shown in uppercase) used in this study are illustrated in Block 3.1 of Figure\ref{fig_framework} and summarized below:

\noindent
\small
\begin{itemize}[topsep=0pt, partopsep=0pt, itemsep=0pt, parsep=0pt]
  \item \textbf{Company}: The main entity.
  \item \textbf{Standard Industrial Classification}: Indicates the company’s type of business.
  \item \textbf{IS\_PARTIAL\_OWNER}: Parent–subsidiary relationship between companies.
  \item \textbf{BELONGS\_TO\_INDUSTRY\_OF}: Links a company to its industry classification.
\end{itemize}


\paragraph{Relation Inference.}
After detecting events, we infer a set of \emph{Event--IMPACTS--Company} relations by linking each identified event to the companies mentioned in the same news article, based on relations such as \emph{News--MENTIONS--Event} and \emph{News--MENTIONS--Company}. In addition, we enhance event information by establishing \emph{Events--HAS\_CATEGORY--Event Category} relations in the above news-derived triples. This linkage allows events and their category-related counterparts to be connected with relevant companies, thereby further linking them to their respective industries within the FinKG framework. These additions are illustrated in Block3.2 of Figure \ref{fig_framework} and summarized as follows:

\noindent
\small
\begin{itemize}[topsep=0pt, partopsep=0pt, itemsep=0pt, parsep=0pt]
  \item \textbf{News}: Entities with properties date, title, URL, and body.
  \item \textbf{Company}: Entities with properties name and identifier.
  \item \textbf{Event}: Entities with properties description, date and geo.
  \item \textbf{Event Category}: The leaf categories of our self-defined event taxonomy.
  \item \textbf{HAS\_CATEGORY}: Links an Event to its corresponding category.
  \item \textbf{MENTIONS}: Links a News Entity to the Event Entity or Company Entity.
  \item \textbf{IMPACTS}: Infers the connection from an Event Entity to a Company Entity.
\end{itemize}




\section{Credit Risk Report Generation Architecture}
\label{sec_generation}

\begin{figure*}
    \centering
    \includegraphics[width=\linewidth]{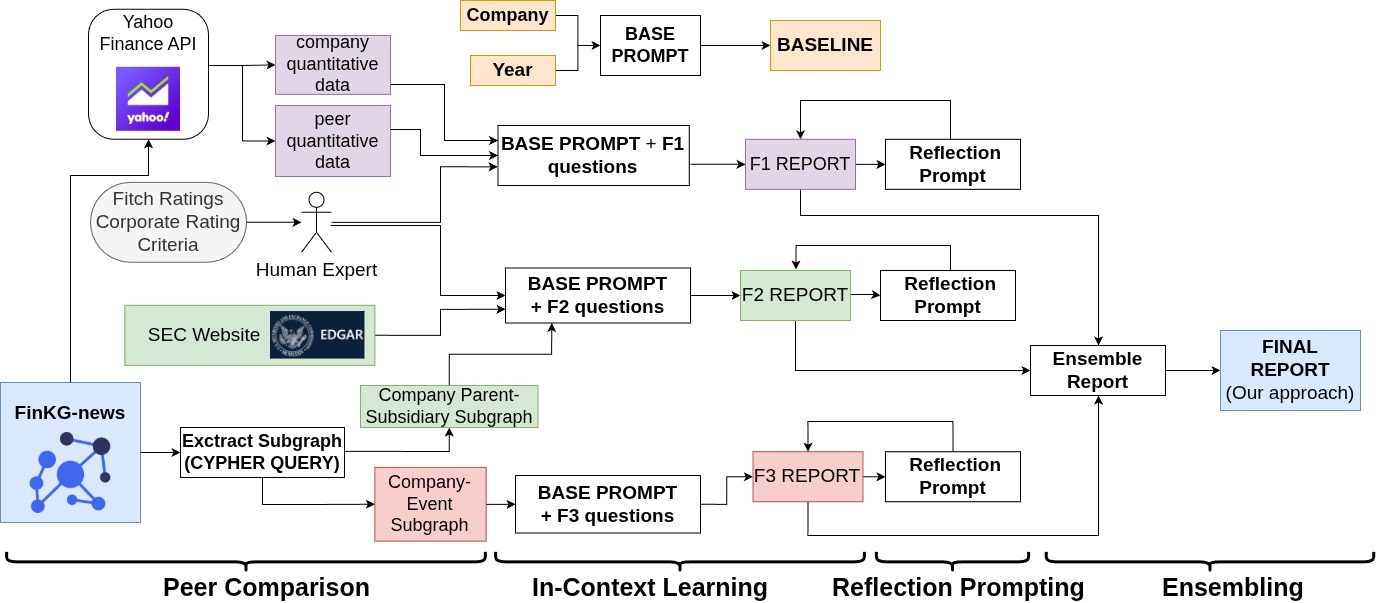}
    \caption{Risk Report Generation Architecture}
    \label{fig:report_generation}
\end{figure*}

To highlight the strength of FinKG-news in reducing hallucinations and effectively identifying important relations, in this section we describe the practical use-case of automatic credit risk report generation. Then, we detail each module of our proposed architecture following the data flow in Figure \ref{fig:report_generation}, including peer comparison between companies for knowledge augmentation, in-context learning for prompt engineering, reflection prompting, and final content ensembling.

Our process starts with constructing the key rating drivers. In the context of a credit risk report, these drivers identify the major reasons and underlying factors behind an assigned credit rating. Following a simplified version of Fitch Ratings’ methodology \cite{fitch_corporate_rating_criteria}, we concentrate on three main categories of key report drivers: 

\begin{itemize}[noitemsep, topsep=0pt]
    \item \textbf{F1-Financial Profile}: Quantitative indicators of company's financial capitalization, profitability, structure and flexibility.
    \item \textbf{F2-Ownership Record}: Internal strategic and organizational characteristics, including competitive positioning and managerial decisions. Here, we also include the ownership and subsidiary structure.
    \item \textbf{F3-Operating Environment}: External macroeconomic, sectoral, regulatory and other external conditions shaping the firm risk context. 
\end{itemize}

\paragraph{Peer Comparison} Peer comparison provides contextual and relative information on a company’s performance and risk profile. Rather than analyzing a company in isolation, our approach focuses on companies within the same industrial classification. For \textit{F1-financial profile}, we contrast key financial ratios and metrics of the target company against one of its peers. For \textit{F3-operating environment}, we consider event impacts not only on the target company but also on other companies within the same sector.

\paragraph{In-Context Learning} Since different drivers capture distinct aspects of a firm's risk profile and rely on different data sources, we apply in-context learning to each driver separately. This allows the LLM to better contextualize the relevant information and generate a dedicated credit risk report for each risk dimension. To further guide the model's reasoning and ensure that its analysis aligns with credit assessment practices, we design a tailored set of guiding questions for each driver category. These question sets are developed with input from a financial expert and they should prevent the LLM from being too generic and superficial, pushing it to carry out an analysis that discovers hidden trends and interesting insights that are difficult for the human eye. 

\paragraph{Reflection Prompting}
Building on prior evidence that LLMs can self-correct by identifying inaccuracies and refining their own outputs \cite{liu2024selfcorrect}, our reflection prompt is designed to revise the initial report. Its goal is to reduce hallucinations, inaccuracies, and irrelevant content, ensure that all key risks are captured and clearly articulated, and remove ambiguous or non-informative statements. 

\paragraph{Ensembling} 
As prior work on chain-of-thought reasoning suggests, decomposing complex analytical tasks into smaller, domain-specific steps can improve both the accuracy and interpretability of LLM-generated outputs \cite{wei2022chainofthought} . In our setting, we have produced a separate report for each dimension of the key report drivers. We then prompt the LLM to ensemble these three reports into a single output that addresses all key drivers jointly.

\begin{figure}[h]
    \centering
    \includegraphics[width=1\linewidth]{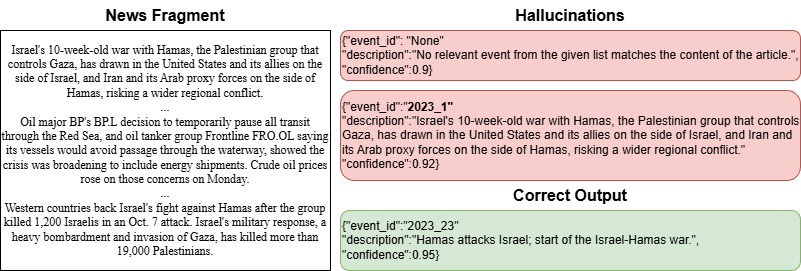}
    \caption{Hallucination examples. The first examples in red color misses a relevant event, and the second fabricates an event with a plausible but wrong identifier. The final example in green color correctly identifies the event and its identifier.}
    \label{fig:hallucination_examples}
\end{figure}

\section{Experiment and Results}
\label{sec_exp}

\subsection{Events Dataset}
\label{subsec:event_dataset}
To construct a representative set of contemporary historical events, we extract the events from the \emph{Wikipedia Timeline of the 21st Century} \cite{wikipedia_timeline_21st_century}, which provides a diverse collection of 354 major global events spanning from 2000 to 2025. Each event already includes a brief description and a date or date range.

Both the geographical annotations and the event categories are generated using a LLM. For each event, we prompt \texttt{GPT-4.1} to independently infer its geographical attributes and its hierarchical category labels. To evaluate the quality of these classifications, we prompt a second, independent LLM (\texttt{gemini-2.5-flash}) to rate the quality of each annotated event from 0 to 1, with a brief reasoning on the score. We then used these ratings to iteratively refine the taxonomy, manually reorganizing, or introducing categories whenever misclassifications were detected, until the taxonomy ensured a complete coverage of all events. The resulting taxonomy is detailed in Appendix \ref{app:taxonomy}.

\subsection{News Dataset}
\label{subsec:data_cleaning}

For our news corpus, we use the FNSPID dataset \cite{dong2024fnspidcomprehensivefinancialnews}, which comprises 15.7 million time-aligned financial news records covering 4,775 companies from 1999 to 2023. Each record contains the full article text and associated company ticker symbols, enabling a mapping between news content and specific companies. We apply a pre-processing pipeline, described in detail in Appendix~\ref{app:news_preprocessing}, which consists of basic data cleaning steps followed by a fast keyword-based filtering procedure. This filtering step uses fuzzy matching to retain only news articles that are likely to mention historical events. After these pre-processing steps, we obtain a clean, event-related financial news dataset comprising approximately 80{,}000 articles.

\subsection{Event Detection from News}
We perform event detection using \texttt{Llama3:70B} model with an event chunk size of N=20. To reduce unnecessary computation, we only consider events from the same news year or earlier, since future events cannot influence past news articles. We set a maximum input length of 2,600 tokens for the model so that if a news article exceeds this limit, it is truncated. This prevents the LLM from ignoring the main task specified in the system prompt.

\paragraph{Post-processing}
After obtaining the LLM-generated outputs, we perform a series of post-processing operations to ensure reliability. We first clean the responses by extracting only the structured list of event detections. Next, we apply confidence-based filtering, discarding events in which the confidence assigned by the LLM is lower than 0.9.

\paragraph{Hallucination Checking}
Finally, we identify hallucinations by self-defined heuristic rules. This involves removing any malformed outputs that don't follow the instructions from the prompt, events without unique identifiers, fabricated identifiers or descriptions, or any references to events not present in the predefined list. We provide some hallucination examples in Figure \ref{fig:hallucination_examples}.

\subsection{FinKG-News Statistics}

In this section we provide an overview of the FinKG-News knowledge graph, including basic statistics and entity/relationship distributions. The fully populated graph comprises 80,418 entities and 261,168 triples, including additional relations inherited from the original FinKG framework that are not explicitly mentioned.  Tables~\ref{tab:entity_counts} and~\ref{tab:relationship_counts} summarize only the counts of the entity and relationship types included in this work, respectively, which account for 15,689 entities and 13,136 triples. Note that the number of news nodes is smaller than the total number of articles in the original news dataset, since only articles that contained events are included.

\begin{table}[t]
\centering
\noindent
\begin{minipage}[c]{0.47\textwidth}
\centering
\caption{Entities in FinKG-News.}
\label{tab:entity_counts}
\footnotesize
\begin{tabular}{lc}
\toprule
\textbf{Entity Type} & \textbf{Count} \\
\midrule
Company                         & 12,407 \\
News                            & 2,428  \\
Std. Industrial Classification & 444  \\
Event                           & 354    \\
Event Category                  & 56     \\
\bottomrule
\end{tabular}
\end{minipage}\hspace{0.03\textwidth}%
\begin{minipage}[c]{0.50\textwidth}
\centering
\caption{Relationships in FinKG-News.}
\label{tab:relationship_counts}
\footnotesize
\begin{tabular}{lc}
\toprule
\textbf{Relationship Type} & \textbf{Count} \\
\midrule
MENTIONS                  & 6,874 \\
IMPACTS                   & 2,391 \\
IS\_PARTIAL\_OWNER\_OF     & 1,897 \\
BELONGS\_TO\_INDUSTRY\_OF  & 1,403 \\
EVENT\_HAS\_CATEGORY      & 571   \\
\bottomrule
\end{tabular}
\end{minipage}
\end{table}



\subsection{F1 Report: Financial Profile}

For the Financial Profile report, we rely on a subset of quantitative metrics referenced in \cite{fitch_corporate_rating_criteria}, limited only to those that can be obtained from publicly available sources such as Yahoo Finance. These include indicators such as \emph{EBITDA}, \emph{total debt}, \emph{free cash flow}, and other \emph{balance sheet}, \emph{income}, and \emph{cash flow statement} measures. For the target company, we extract data for the reporting year and the three previous years. The resulting metrics are organized into a table and injected into the prompt.

We complement the company-level data with a simplified peer comparison. Specifically, we sample ten company entities from FinKG-News belonging to the same industry category as the target firm. From this subset, we select as peer the company whose total assets are most similar to those of the target. We then extract and format the same four-year financial history for the peer company and include it alongside the target firm’s data in the prompt.

The guiding questions provided to the LLM aim to evaluate key dimensions of financial health, including \emph{profitability}, \emph{liquidity}, \emph{leverage}, \emph{coverage ratios}, and \emph{cash flow generation}. 

\begin{figure}[h]
    \centering
    \includegraphics[width=0.9\linewidth]{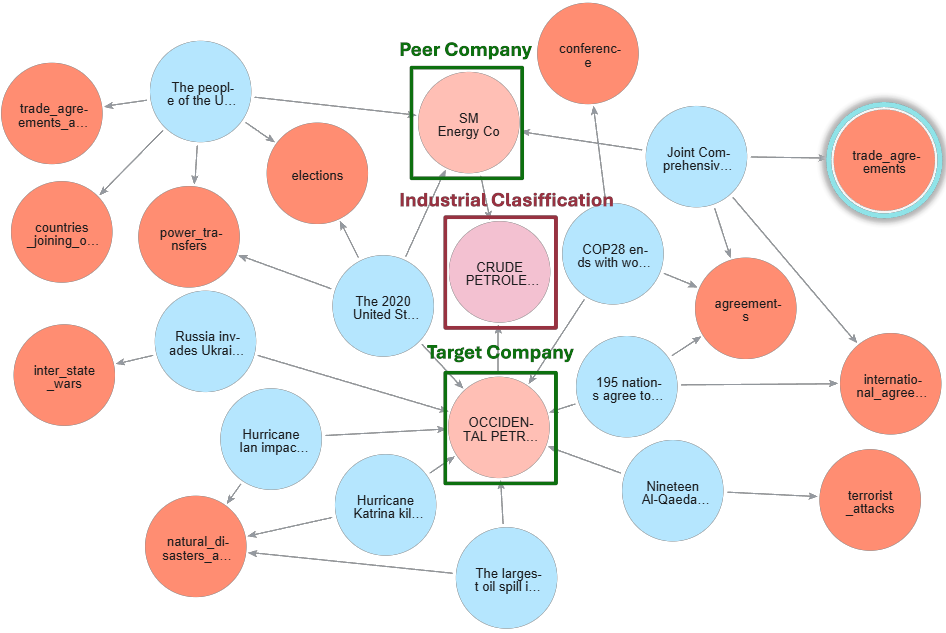}
    \caption{Sample of event subgraph for Occidental Petroleum Corp., including one sector peer (SM Energy Co.) from the Crude Petroleum \& Natural Gas industry. Nodes are colored by type: companies in light orange, events in blue, the sector node in pink, and event categories in orange.}
    \label{fig:subgraph_example}
\end{figure}

\subsection{F2 report: Ownership Records}
For the ownership records, we focus mainly on internal factors and the company’s subsidiary--parent structure. We analyze the owner and issuer transaction tables available on the SEC website. Using web scraping, we extract all transactions corresponding to the report’s target year and inject them into the LLM prompt. This serves as a primary data source because insider activity may reveal evidence of governance risks, the level of confidence of senior executives and major stakeholders, and other risks related to potential misalignment or strategic issues. For instance, large volumes of insider selling may sometimes indicate a lack of confidence in the company.
If the company has a parent or a subsidiary, we include this parent-subsidiary structure in the analysis. To incorporate this, we query FinKG-News for \texttt{IS\_PARTIAL\_OWNER} relations in both directions and format the resulting triples in a way that is easy for the LLM to interpret.

\subsection{F3 report: Operating Environment}

To describe risks related to the external environment, we rely on the events stored in FinKG-News. We query FinKG-News to generate a subgraph centred on the target company, including: (i) all events that have affected the company; (ii) the Standard Industrial Classification to which the company belongs; (iii) other companies within the same Industrial Classification that have associated events; and (iv) the event categories associated with all these nodes. The resulting subgraph is then formatted into a structured, LLM-friendly representation.

By including the company's full history of events, we can identify vulnerabilities that have recurred over time. Moreover, extending the subgraph to include sector peers highlights patterns of exposure. For instance, from Figure~\ref{fig:subgraph_example}, we can deduce that companies in the oil sector (pink node) are particularly sensitive to events related to climate and environmental policies or armed conflicts such as the Russia–Ukraine war. 

\section{Evaluation}
\label{sec_eval}

\subsection{Evaluation Matrix} 

We generate baseline reports by prompting the LLM with only the core task instructions and the descriptive definitions of the three key rating driver categories. We then benchmark the baseline and our work using evaluations from financial experts and an automatic scoring framework. We evaluate reports along two categories: content quality and hallucination detection.
All scores are normalized to the unit interval. Section \ref{sec:quaneval} reports the quantitative evaluations of our method over the baseline at both the aggregate and individual firm cases. Section \ref{sec:qualeval} records the qualitative assessments from financial experts after reviewing the reports.

\subsection{Quantitative Evaluation} 
\label{sec:quaneval}

We select five Small and Medium Enterprises (SMEs). Table \ref{tab:total_scores} shows that, under both human and automatic evaluations, our method delivers systematic improvements in overall scores rather than firm-specific performance. Based on expert ratings, the average improvement relative to the baseline is about 18\%-34\%, while the largest improvement under automatic scoring reaches 16.5\%. Table \ref{tab:indicator_scores} reports indicator-level results for the firm with the largest expert-based evaluation uplift(+34\%), suggesting that the improvement are not driven by simple stylistic polishing but by a shift toward traceable data analysis. In the hallucination detection, experts report significant increases in \textit{Attention} (+70\%), \textit{Correctness} (+65\%), and \textit{Fabrication} (+65\%). In the content quality, \textit{Usefulness} improves by 65\%. Meanwhile, \textit{Balance} (-5\%) and \textit{Eco-Validity} (-20\%) decline modestly, consistent with a trade-off whereby more explicit, conditional predictions sometimes replace overly conservative or tautological statements. Notably, for the hallucination-related agreement, the automatic evaluator still remains less sensitive to hallucinations than expert review. Especially for \textit{Fabrication} detection, the automated methods fail to identify baseline fabrications as reliably as human experts.

\begin{table*}[t]
\centering
\caption{Total Scores Across Companies}
\label{tab:total_scores}
\footnotesize
\setlength{\tabcolsep}{1pt}
\renewcommand{\arraystretch}{1.0}

\begin{tabular}{lccc c ccc c}
\toprule
& \multicolumn{3}{c}{Human Eval.} & & \multicolumn{3}{c}{Automatic Eval.} & Agreement of \\
\cmidrule(lr){2-4}\cmidrule(lr){6-8}
Company & Base. & Ours & Improv. && Base. & Ours & Improv. & Improv. \\
\midrule
\textit{Alaska Air}
& 0.653 & 0.838 & 0.185
&& \underline{0.735} & 0.750 & 0.015 & 0.170 \\

\textit{Core Civics}
& 0.500 & 0.840 & \textbf{0.340}
&& 0.695 & 0.800 & 0.105 & \textbf{0.235} \\

\textit{Disney}
& \underline{0.659} & 0.852 & 0.193
&& 0.705 & \textbf{0.870} & \textbf{0.165} & 0.028 \\

\textit{Merck and Co}
& \textbf{0.668} & \textbf{0.863} & 0.195
&& \textbf{0.745} & 0.745 & 0.000 & 0.195 \\

\textit{Occidental Petrol.}
& 0.536 & \underline{0.861} & \underline{0.325}
&& 0.720 & \underline{0.841} & \underline{0.121} & \underline{0.204} \\
\bottomrule
\end{tabular}
\end{table*}

\begin{table*}[t]
\centering
\caption{Evaluation for \textit{Core Civics}}
\label{tab:indicator_scores}
\footnotesize
\setlength{\tabcolsep}{1pt}
\renewcommand{\arraystretch}{1.0}

\begin{tabular}{@{}m{1.7cm}lccc c ccc c@{}}
\toprule
& & \multicolumn{3}{c}{Human Eval.} && \multicolumn{3}{c}{Automatic Eval.} & Agreement of \\
\cmidrule(lr){3-5}\cmidrule(lr){7-9}
Category & Indicator & Base. & Ours & Improv. && Base. & Ours & Improv. & Improv. \\
\midrule

\multirow{5}{=}{Content\\Quality}
& Balance          & \underline{0.75} & 0.70 & -0.05 && 0.80 & 0.75 & -0.05 & 0.00 \\
& Coherence        & \underline{0.75} & \textbf{0.95} & 0.20 && 0.80 & \underline{0.85} & 0.05 & 0.15 \\
& Convincing       & 0.40 & 0.80 & 0.40 && 0.45 & \textbf{0.90} & \textbf{0.45} & 0.05 \\
& Informativeness  & 0.60 & 0.70 & 0.10 && 0.60 & \textbf{0.90} & 0.30 & 0.20 \\
& Usefulness       & 0.25 & \underline{0.90} & \underline{0.65} && 0.50 & \textbf{0.90} & \underline{0.40} & 0.25 \\
\midrule

\multirow{5}{=}{Halluc.\\Detection}
& Attention        & 0.25 & \textbf{0.95} & \textbf{0.70} && 0.60 & \textbf{0.90} & 0.30 & 0.40 \\
& Correctness      & 0.25 & \underline{0.90} & \underline{0.65} && 0.60 & 0.80 & 0.20 & \underline{0.45} \\
& Ecovalidity      & \textbf{1.00} & 0.80 & -0.20 && \underline{0.85} & 0.70 & -0.15 & 0.05 \\
& Emotion          & 0.50 & 0.80 & 0.30 && \textbf{0.95} & \underline{0.85} & -0.10 & 0.40 \\
& Fabrication      & 0.25 & \underline{0.90} & \underline{0.65} && 0.80 & 0.45 & -0.35 & \textbf{1.00} \\
\bottomrule
\end{tabular}
\end{table*}

\subsection{Qualitative Evaluation} 
\label{sec:qualeval}

Qualitative feedback from financial experts highlights three main improvements. First, our method introduces peer benchmarks (industry averages and large-firm references), putting the change of financial indicators in comparative context and making risk assessments more objective. In contrast, baseline reports lack such context and tend to rely on subjective or emotionally charged descriptions that may amplify investor panic. Second, our approach integrates external information such as news, insider trading signals, and corporate governance indicators. Experts note that these dimensions are highly relevant to investors but are largely absent from baseline analyses, which rely on generic risk narratives applicable to most firms. Third, experts indicate substantially fewer hallucinations in our reports. By integrating multiple information sources and a financial knowledge graph, our conclusions are supported by explicit reasoning chains, making the reports more convincing than the baseline. 
For example, The baseline report equates ``declining profitability'' with ``higher refinancing risk,'' but declining profitability must coincide with other conditions (e.g., rising leverage and tighter credit) to raise refinancing risk. In contrast, our reports present the full logic explicitly.

\subsection{Event Detection Evaluation}

We evaluate the event detection component on a manually annotated set of 100 news articles, framing the task as a multi-label classification problem over 354 possible event classes. Overall performance is reported using standard micro- and macro-averaged precision, recall, and F1-score, as well as Exact Match Accuracy. Table~\ref{tab:results} (Appendix~\ref{app:event_detection_eval}) reports the full evaluation results. The model achieves a micro F1-score of 0.72 and an Exact Match Accuracy of 0.64 on the manually annotated test set. 

\section{Related Work}
\label{sec_relatedw}

\subsection{Financial Knowledge Graphs}

Knowledge Graphs represent real-world knowledge in a structured and machine-interpretable way, typically through triples of the form (subject, predicate, object) \cite{knowledgegraphs}. Several large-scale, general-purpose knowledge graphs have been developed to capture broad, cross-domain world knowledge, including DBpedia, Wikidata, YAGO, and Freebase. Building on this, subsequent work has focused on the construction of domain-specific knowledge graphs. In the financial domain, financial knowledge graphs (FKGs) have been proposed to support a wide range of applications, including information retrieval and financial question answering \cite{finkg}, stock price prediction \cite{TAO20224322}, fraud detection \cite{WEN2022773}, portfolio management \cite{findkg}, and risk management \cite{liu2024risk}. These graphs typically model companies, financial instruments, executives, sectors, and inter-company relations.

One of the earliest efforts toward a financial knowledge graph is presented in \cite{first_finkg}, where the authors construct a graph by extracting entities and relations from heterogeneous sources such as unstructured text, web pages, and data feeds. 
Later studies also build knowledge graphs from textual sources such as SEC 10-K filings \cite{FinReflectKG,xu2025fincakg} and financial news articles \cite{elhammadi2020high}. However, not all financial knowledge graphs rely on text-based extraction. For instance, FinKG \cite{finkg} prioritizes high-quality structured data sources, such as information from the SEC website, to construct a reliable financial knowledge graph with a well-defined ontology. Our work builds upon this line of research by extending FinKG with event-centric information extracted from financial news. 

With the rise of large language models and their strong information extraction capabilities, recent work has shifted toward prompt-based knowledge graph construction. In this setting, an LLM is typically guided by a predefined schema specified in the prompt and tasked with extracting entities and relations accordingly \cite{findkg,FinReflectKG,chen-etal-2024-knowledge}. FINDKG \cite{findkg} constructs a dynamic financial knowledge graph from financial news using an LLM-based generator with a predefined schema, while FinReflectKG \cite{FinReflectKG} presents a large-scale graph built from SEC 10-K filings. Specific events and companies have to be implicitly inferred from text, making entity coverage and semantic consistency highly dependent on the model. In contrast, our approach specifies concrete and sometimes less well-known events and companies, particularly smaller-cap firms that are more likely to benefit from this approach, resulting in a richer and more diverse graph.


\subsection{LLMs}

A crucial aspect of leveraging LLMs effectively is prompt engineering. Carefully crafted inputs that specify the context and desired output, guide LLMs toward producing relevant and task-specific results \cite{schulhoff2024prompt}. Advanced strategies, such as chain-of-thought prompting, can decompose complex tasks into smaller reasoning steps, improving consistency and interpretability \cite{wei2022chainofthought}. Our approach adopts this principle of task decomposition.



Generating text in the financial domain presents unique challenges due to highly specialized terminology and the strict requirement for factual accuracy, particularly in applications such as credit risk assessment and financial reporting.
To address these challenges, several domain-specific LLMs have been developed. Models such as FinGPT \cite{yang2023fingpt}, BloombergGPT \cite{wu2023bloomberggpt}, and FinBERT \cite{liu2021finbert} are pre-trained or fine-tuned on large financial corpora, enabling better understanding of domain-specific terminology and financial reasoning. However, some models like BloombergGPT are closed-source, limiting accessibility and reproducibility. Moreover, recent studies have shown that large general-purpose models, such as GPT-4, can consistently outperform specialized fine-tuned models on niche financial tasks \cite{li-etal-2023-chatgpt}, especially when combined with appropriate techniques, including careful prompt engineering \cite{maharjan2024openmedlm} and retrieval-augmented generation \cite{lewis2020retrieval}. These approaches that incorporate external knowledge into prompts help to mitigate hallucinations and improve output fidelity. 


Only limited work focuses specifically on financial report generation. Chen et al.~\cite{chen-etal-2024-knowledge} introduces a knowledge-graph-grounded retrieval-augmented framework for financial market analysis report generation, reducing hallucinations and improving logical consistency.




\section{Conclusions}
\label{sec_concl}

In this work, we introduced FinKG-News, a company-centric and event-aware financial knowledge graph constructed automatically from news. Using this knowledge graph, we designed a credit risk report generation framework based on in-context learning. Our results show higher report quality and fewer hallucinations compared to the baseline. Evaluation further highlights that hallucination detection and quality assessment remain challenging for automatic methods, highlighting the importance of human expert evaluation. In brief, our system is capable to generate reliable, evidence-based credit risk reports for even small- and mid-cap companies, a setting in which current LLMs frequently hallucinate due to the sparse reference data.

\section*{Limitations}
\label{sec_limt}

Due to the context window size of some LLMs, we introduce the chunking module for event detection. This solution addresses the context window limitation by reducing the size of the system prompt; however, it introduces additional limitations related to computation time and raises concerns about the scalability of the approach. Another limitation appears when the kind of impact relation between an event and a company is not evident and requires additional context. In such cases, an LLM may generate reports with more generic or ambiguous explanations. This opens possible directions for future work, such as enriching Event-Company relations with causal or contextual evidence extracted directly from the same news articles or implementing a causality check module during triple inference.

\begin{credits}

\subsubsection{\discintname}
The authors have no competing interests to declare that are
relevant to the content of this article.
\end{credits}
%
%
%
\bibliographystyle{splncs04}
\bibliography{biblio}
%
\appendix
\FloatBarrier

\section{News Dataset Pre-Processing Details}
\label{app:news_preprocessing}

\begin{figure}[h]
    \centering
    \includegraphics[width=0.85\linewidth]{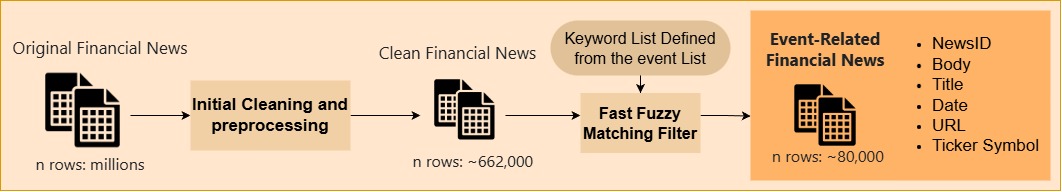}
    \caption{News Dataset Preprocessing}
    \label{fig:news_preprocessing}
\end{figure}

In this section, we provide a detailed description of the procedures carried out for cleaning and pre-processing our News Dataset. The news pre-processing pipeline is depicted in Figure \ref{fig:news_preprocessing}.

\vspace{-6pt}
\vspace{-15pt}

\paragraph{Initial Cleaning and Preprocessing.}
We perform an initial preprocessing stage to ensure data consistency and relevance. First, we remove all entries with missing values and retain only records containing the following fields: \textit{date}, \textit{article title}, \textit{article body}, \textit{ticker symbol}, and \textit{URL}. Second, we reduce the dataset to news related to companies included in the S\&P 600, S\&P 500, or S\&P 400 indices.

\paragraph{Fast Fuzzy Matching Filter}
Since many articles in the news corpus do not reference any historical events, we introduce an additional filtering step to identify articles that may potentially mention such events. To this end, we construct a keyword list derived from the set of historical events generated in Section~\ref{subsec:event_dataset}. We then apply a fast fuzzy matching procedure to retain only those news articles that match at least one event-related keyword. Specifically, we use the \texttt{partial\_ratio} scorer from the \texttt{RapidFuzz} Python library \cite{rapidfuzz} with a threshold of 0.95. This function measures similarity between a short string (the keyword) and a longer string (the news article) by matching the shorter string against the best-aligned substring of the longer text. This step reduces the size of the dataset while preserving articles that are most likely to mention historical events.

\section{Self-Defined Event Category Taxonomy}
\label{app:taxonomy}

\vspace{-15pt}%
\newcolumntype{L}{>{\raggedright\arraybackslash}p{1.0cm}}  
\newcolumntype{M}{>{\raggedright\arraybackslash}p{3.0cm}}  
\begin{table*}[h]
\centering
\caption{Event Category Taxonomy}
\label{tab:taxonomy}
\scriptsize
\setlength{\tabcolsep}{2.5pt}
\renewcommand{\arraystretch}{1.05}
\begin{tabularx}{\textwidth}{@{}>{\raggedright\arraybackslash}p{0.16\textwidth}
                              >{\raggedright\arraybackslash}p{0.24\textwidth}
                              X@{}}
\toprule
\textbf{Level 1} & \textbf{Level 2} & \textbf{Level 3} \\
\midrule

\textbf{POLITICAL} & Leadership Changes
& Elections; Power Transfers; Leader Deaths/Assassinations; Coup d'État. \\

& Wars and Armed Conflicts
& Inter-state Wars; Civil Wars; Border Conflicts; Terrorist Attacks;
  Military Escalations; Peace Agreements. \\

& Economy
& Recessions and Crises; Trade Agreements and Blocs; Financial Scandals;
  Currency and Monetary Changes. \\

& Climate and Environment Policies
& Agreements; Reports; Conferences. \\

& Legal Milestones
& Constitutions; New Legislations; Other. \\

& Social Movements
& Protests and Riot-triggering Events; Revolutions and Uprisings;
  Ethnic and Religious Tensions Escalating. \\

& Diplomatic Relations
& International Agreements; Countries Joining Organizations and Blocs;
  Important Inter-country Meetings. \\

\midrule

\textbf{NON-POLITICAL} & Non-political Disasters and Crises
& Natural Disasters and Environmental Events; Health Crises and Pandemics;
  Other Disasters or Tragedies. \\

& Technology or Science Breakthroughs
& Space Exploration and Astronomy; Medical and Biological Discoveries;
  Disruptive Technologies; Consumer Tech Milestones; Other Inventions and
  Discoveries. \\

& Cultural Events and Influences
& Major Sports and Arts Events (Olympics, Concerts); Iconic Deaths and Births;
  Religion-related; Artistic and Cultural Releases; Internet and Social
  Network Trends. \\

& Corporate and Market
& Founding of Major Companies; Bankruptcy or Scandal of Influential Firms;
  Mergers and Acquisitions with Global Impact; Other market milestones. \\

\bottomrule
\end{tabularx}
\end{table*}

\section{Event Detection Evaluation}
\label{app:event_detection_eval}






\vspace{-20pt}%
\begin{table}[h]
\centering
\begin{tabular}{lccc}
\hline
 & Precision & Recall & F1-score \\
\hline
Micro & 0.6941 & 0.7564 & 0.7239 \\
Macro & 0.5480 & 0.5223 & 0.5235 \\
\hline
\multicolumn{3}{l}{Exact Match Accuracy} & 0.6364 \\
\hline
\end{tabular}
\caption{Event Detection Evaluation Results}
\label{tab:results}
\end{table}

\vspace{-10pt}%
Table~\ref{tab:results} summarizes the event detection results on the manually annotated test set. Micro-averaged scores are consistently higher than macro-averaged scores, indicating uneven performance across event classes. This behavior may partly reflect imbalance in the news dataset, where some events occur more frequently than others and also exhibit stronger detection performance. 




\clearpage

\end{document}